\documentclass[letterpaper]{article} % DO NOT CHANGE THIS
\usepackage{aaai2026}  % DO NOT CHANGE THIS
\usepackage{times}  % DO NOT CHANGE THIS
\usepackage{helvet}  % DO NOT CHANGE THIS
\usepackage{courier}  % DO NOT CHANGE THIS
\usepackage[hyphens]{url}  % DO NOT CHANGE THIS
\usepackage{graphicx} % DO NOT CHANGE THIS
\usepackage{natbib}  % DO NOT CHANGE THIS AND DO NOT ADD ANY OPTIONS TO IT
\usepackage{caption} % DO NOT CHANGE THIS AND DO NOT ADD ANY OPTIONS TO IT
\usepackage{algorithm}
\usepackage{algorithmic}

\usepackage{newfloat}
\usepackage{listings}
\usepackage{multirow}
\DeclareCaptionStyle{ruled}{labelfont=normalfont,labelsep=colon,strut=off} % DO NOT CHANGE THIS
\floatstyle{ruled}
\newfloat{listing}{tb}{lst}{}
\floatname{listing}{Listing}
\title{SCIR: A Self-Correcting Iterative Refinement Framework for Enhanced Information Extraction Based on Schema}

\author{
    Yushen Fang,
    Jianjun Li\thanks{Corresponding author.},
    Mingqian Ding,
    Chang Liu,
    Xinchi Zou,
    Wenqi Yang
}
\affiliations{
     School of Computer Science and Technology, Huazhong University of Science and Technology \\
    \{yushen\_fang, jianjunli, mingqianding, diu, xinchizou, yangwenqi\}@hust.edu.cn \\
}

\begin{document}
	\maketitle
	
	\begin{abstract}
		
		Although Large language Model (LLM)-powered  information extraction (IE) systems have shown impressive capabilities, current fine-tuning paradigms face two major limitations: high training costs and difficulties in aligning with LLM preferences. To address these issues, we propose a novel universal IE paradigm—the Self-Correcting Iterative Refinement (SCIR) framework—along with a Multi-task Bilingual (Chinese-English) Self-Correcting (MBSC) dataset containing over 100,000 entries. The SCIR framework achieves plug-and-play compatibility with existing LLMs and IE systems through its Dual-Path Self-Correcting module and feedback-driven optimization, thereby significantly reducing training costs. Concurrently, the MBSC dataset tackles the challenge of preference alignment by indirectly distilling GPT-4's capabilities into IE result detection models.	Experimental results demonstrate that SCIR outperforms state-of-the-art IE methods across three key tasks— named entity recognition, relation extraction, and event extraction—achieving a 5.27 percent average improvement in span-based Micro-F1 while reducing training costs by 87 percent compared to baseline approaches. These advancements not only enhance the flexibility and accuracy of IE systems but also pave the way for lightweight and efficient IE paradigms.
	\end{abstract}

	\begin{links}
		\link{Code \& Datasets \& Extended version}{https://github.com/Sheehan-Fang/SCIR}
	\end{links}
	
	\section{Introduction}
	\label{sec:introduction}
	% !TEX spellcheck = en_US
% !TeX root = main.tex

\begin{figure}[t]
	\centering
	\includegraphics[width=\linewidth]{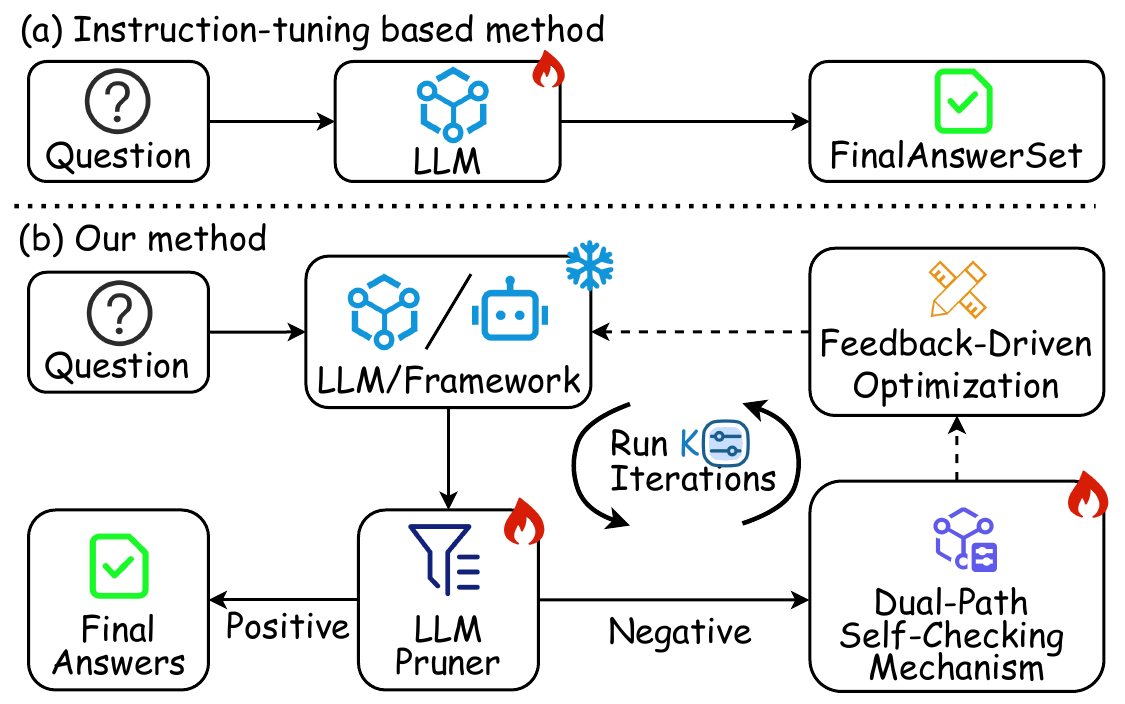}
	\caption{Traditional IE methods vs. our SCIR framework.}
	\label{fig:sample}
\end{figure}

%Information Extraction (IE), as a core technology in natural language processing, focuses on automatically extracting and structuring key information from unstructured text~\cite{DBLP:conf/scie/Wilks97}. Its research scope primarily encompasses three fundamental tasks: named entity recognition (NER), relation extraction (RE), and event extraction (EE). Recent years have witnessed remarkable advancements in large language models (LLMs), demonstrating their exceptional generalization capabilities and versatility across various downstream tasks, particularly in the IE domain. As LLMs gradually become mainstream solutions in natural language processing tasks~\cite{DBLP:journals/corr/abs-2312-00752,DBLP:conf/aaai/JiangCXLG25,DBLP:conf/iclr/0001Z0S23,DBLP:conf/coling/Xiong0XC25,DBLP:conf/acl/ZhangG023,DBLP:conf/iclr/WanH0QB024,DBLP:journals/corr/abs-2309-13064,DBLP:journals/fcsc/MaoGFXMHG25}, the exploration of their capability boundaries in IE has garnered increasing academic attention.

Information Extraction (IE) stands as a pivotal technology within the field of Natural Language Processing (NLP), dedicated to automatically extracting and structure key information from unstructured text~\cite{DBLP:conf/scie/Wilks97}. Its research primarily revolves around three fundamental tasks: named entity recognition (NER), relation extraction (RE), and event extraction (EE). In recent years, as LLMs increasingly establish themselves as mainstream solutions for NLP tasks~\cite{DBLP:journals/corr/abs-2312-00752,DBLP:conf/aaai/JiangCXLG25,DBLP:conf/iclr/0001Z0S23,DBLP:conf/coling/Xiong0XC25,DBLP:conf/iclr/WanH0QB024,DBLP:journals/corr/abs-2309-13064,DBLP:journals/corr/abs-2505-11739,DBLP:journals/fcsc/MaoGFXMHG25}, 
there has been a burgeoning academic interest in exploring the full potential and boundaries of LLMs within the IE domain.
Several notable advancements have emerged in this pursuit~\cite{DBLP:journals/corr/abs-2312-15548,DBLP:conf/acl/GuiYYZSLC24,DBLP:journals/corr/abs-2412-20005,chunkuie2025}. For instance, OneKE~\cite{DBLP:journals/corr/abs-2412-20005} achieved significant performance enhancements by leveraging data synthesis and fine-tuning strategies tailored to IE tasks.  ChunkUIE~\cite{chunkuie2025} introduced an innovative chunk-based extraction methodology, offering a fresh perspective on structured information retrieval from text. Furthermore, RUIE~\cite{DBLP:conf/coling/LiaoDH025} integrated retrieval augmentation techniques into IE workflows, setting a new state-of-the-art benchmark.

Despite their promise, LLM-based IE models still face two major challenges: (1)~\textbf{High training costs and limited model flexibility.} 
%Current mainstream methods typically rely on fine-tuning techniques to enhance domain-specific performance, but this not only consumes substantial computational resources and time but also weakens the model's semantic understanding capabilities and limits its generalization performance in new domains. More critically, existing frameworks are often tightly coupled with specific underlying models, making it difficult to adapt to their rapid iteration cycles (e.g., the GPT series updates every 3-6 months). The high cost of retraining (often taking weeks or even months) further hinders users from promptly adopting newer and more advanced models. 
Current mainstream methods predominantly rely on fine-tuning techniques to enhance domain-specific performance. However, this approach not only demands substantial computational resources and time, but also weakens the model's semantic understanding capabilities and restricts its generalization performance in new domains. More critically, existing frameworks are often tightly coupled with specific underlying models, making it difficult to adapt to their rapid iteration cycles (e.g., the GPT series updates every 3-6 months). The high cost of retraining, often taking weeks or even months, further impedes users from promptly adopting newer and more advanced models.
(2)~\textbf{Difficulty in aligning model preferences}. 
%Existing information extraction models are largely constrained by the inherent biases and blind spots in human annotations. For edge cases that are frequently overlooked or prone to errors during annotation, models lack remedial mechanisms and cannot fundamentally eliminate such errors simply by increasing data volume. Moreover, traditional supervised training follows a static annotation-to-static inference paradigm, lacking dynamic feedback and self-correction capabilities for model error patterns, which hinders effective improvement in output accuracy and consistency when facing unknown or complex contexts.
Existing information extraction models are significantly constrained by the inherent biases and blind spots present in human annotations. For edge cases that are often overlooked or prone to errors during the annotation process, these models lack remedial mechanisms and cannot completely eliminate such errors simply by increasing the volume of data. Furthermore, traditional supervised training adheres to a ``static annotation to static inference" paradigm, lacking dynamic feedback and self-correction capabilities for addressing model error patterns. This hinders the models' ability to effectively enhance the accuracy and consistency of their outputs when confronted with unknown or complex contexts.

To address these challenges, we propose a novel IE paradigm (as depicted in Figure~\ref{fig:sample}) and introduce the Self-Correcting Iterative Refinement (SCIR) framework, which achieves breakthroughs through the following innovations. Specifically, to  tackle the first challenge, we designed the SCIR framework to eliminate the need for fine-tuning extraction models. This framework leverages a Dual-Path Self-Correcting mechanism and a Feedback-Driven Optimization mechanism to enhance the flexibility of IE systems.  The Dual-Path Self-Correcting mechanism verifies the completeness of extraction results through two pathways: redundancy detection and missing detection. The Feedback-Driven Optimization mechanism generates iterative prompts based on verification results to drive a context-learning-based iterative generation process. This design enables flexible substitution of IE models while requiring only a single training session for the Dual-Path Self-Correcting mechanism, regardless of model replacements—significantly enhancing system flexibility.
%The Dual-Path Self-Correcting mechanism verifies the completeness of extraction results through two pathways: redundancy detection and missing detection. The Feedback-Driven Optimization mechanism generates iterative prompts based on verification results to drive a context-learning-based iterative generation process. This design enables flexible substitution of IE LLMs while requiring only a single training session for the Dual-Path Self-Correcting mechanism, regardless of model replacements—significantly enhancing system flexibility. 
To address the second challenge, we constructed a Multi-task Bilingual Self-Correcting (MBSC) training set based on the IEPile dataset~\cite{DBLP:conf/acl/GuiYYZSLC24}, specifically designed for model preference alignment training. Unlike traditional static datasets reliant on manual annotations, the MBSC dataset centers on error instances generated by GPT-4 in information extraction tasks. It systematically collects edge cases often overlooked, annotation blind spots, and model error-prone points, followed by multi-task labeling. By incorporating real-world error scenarios, MBSC enhances the diversity of training samples, enabling models trained on MBSC to identify biases in extraction results and provide dynamic feedback signals to extraction models. Our main contributions can be summarized as follows:

%We conducted zero-shot transfer evaluations on 11 benchmarks covering relation extraction, named entity recognition, and event extraction, with SCIR achieving an absolute percentage increase of 5.27\% in the average F1-score. Additionally, compared to full-parameter fine-tuning, SCIR's plug-and-play design enables rapid adaptation to various LLMs while significantly reducing training costs. The main contributions of this paper can be summarized as follows:

%We conducted zero-shot transfer evaluation on 11 benchmarks covering relation extraction, named entity recognition, and event extraction, with SCIR achieving an average F1-score improvement of 5.27 percentage points. Additionally, compared to full-parameter fine-tuning, SCIR's plug-and-play design enables rapid adaptation to various LLMs while significantly reducing training costs. The main contributions of this paper can be summarized as follows: 

%\begin{itemize}
%	\item ~\textbf{Dataset Construction}: Constructed MBSC, a training dataset designed for error correction model training and preference alignment.
%	\item ~\textbf{Framework Design}: Proposed the SCIR framework that achieves strong generalization performance with minimal training requirements.
%	\item ~\textbf{Experimental Verification}: Comprehensive experimental verification across 11 multilingual benchmark datasets, demonstrating SCIR's effectiveness.
%\end{itemize}

\begin{itemize}
	\item \textbf{Framework Paradigm Shift}: We propose SCIR, a pioneering fine-tuning-free IE paradigm that achieves exceptional generalization via integrating Dual-Path Self-Correcting and Feedback-Driven Optimization mechanisms, enabling seamless IE base model substitution and iterative refinement while ensuring cost efficiency.
	\item \textbf{Specialized Dataset Synthesis}: We introduce MBSC, an innovative dataset tailored for error correction and preference alignment in IE models, systematically capturing edge cases, annotation blind spots, and model errors to enhance training diversity and robustness.
	\item \textbf{Empirical Performance Breakthrough}: Through comprehensive zero-shot transfer evaluations across 11 multilingual benchmarks, we demonstrate SCIR's outstanding effectiveness with an 5.27\% average F1-score increase, underscoring its potential to revolutionize IE by providing a plug-and-play solution that cuts training costs while maintaining high performance.
\end{itemize}

	\section{Related Work}
	\label{sec:relatedwork}
	% !TEX spellcheck = en_US
% !TeX root = main.tex

%The development of information extraction (IE) has progressed through three key stages. Initially, rule-based systems~\cite{Chiticariu2013RuleBasedIE,Maturana2017DocumentSpanners,Thenmozhi2018RCEOIE} relied on manually crafted patterns (e.g., regular expressions) for domain-specific tasks, but their limited generalizability and high maintenance costs spurred the adoption of statistical and machine learning methods. Approaches like Hidden Markov Models and SVMs~\cite{DBLP:journals/ml/CortesV95} leveraged annotated corpora to improve generalization. Today, deep learning dominates IE, with Transformer-based models like BERT~\cite{DBLP:conf/naacl/DevlinCLT19} achieving state-of-the-art performance through large-scale pretraining, particularly in relation and event extraction.

The evolution of information extraction (IE) has undergone three distinct phases. Early rule-based systems \cite{Chiticariu2013RuleBasedIE,Maturana2017DocumentSpanners,Thenmozhi2018RCEOIE} utilized manually engineered patterns (e.g., regular expressions) for domain-specific tasks, but their poor generalization across domains and high maintenance costs motivated the shift toward statistical and machine learning approaches. Methods such as Hidden Markov Models and Support Vector Machines \cite{DBLP:journals/ml/CortesV95} leveraged annotated corpora to improve model adaptability. Currently, deep learning dominates IE research, with Transformer-based architectures like BERT \cite{DBLP:conf/naacl/DevlinCLT19} achieving state-of-the-art performance through large-scale pretraining, particularly in relation and event extraction tasks.

%Modern IE frameworks comprise two paradigms: open IE and schema-based IE~\cite{DBLP:conf/emnlp/Qi0W00L24}. Open IE extracts unstructured semantic relationships (e.g., for QA) without predefined schemas~\cite{DBLP:conf/aaai/Lou0DJLH0023}, requiring  post-processing (e.g., clustering) for standardization~\cite{DBLP:conf/emnlp/Ma0HS23,DBLP:conf/acl/0001LDXLHSW22}. Schema-based IE, prevalent in vertical domains, follows structured templates (e.g., entity-relation definitions or event hierarchies) to ensure precision and interoperability~\cite{DBLP:conf/naacl/TedeschiN22,DBLP:conf/acl/ZhengWBHZX17}. These schemas range from flat lists to nested structures, guiding extraction with explicit constraints~\cite{DBLP:conf/naacl/WangSLOWZLWG25}.

Modern IE frameworks primarily follow two paradigms: open IE and schema-based IE \cite{DBLP:conf/emnlp/Qi0W00L24}. Open IE systems extract unstructured semantic triples (e.g., for question answering) without predefined schemas \cite{DBLP:conf/aaai/Lou0DJLH0023}, necessitating post-hoc standardization through clustering or alignment techniques \cite{DBLP:conf/emnlp/Ma0HS23,DBLP:conf/acl/0001LDXLHSW22}. In contrast, schema-based IE—common in specialized domains—adheres to structured templates (e.g., entity-relation taxonomies or event hierarchies) to ensure extraction precision and interoperability \cite{DBLP:conf/naacl/TedeschiN22,DBLP:conf/acl/ZhengWBHZX17}. These schemas range from flat entity lists to complex nested structures, providing explicit constraints to guide the extraction process \cite{DBLP:conf/naacl/WangSLOWZLWG25}.

%Recent advances integrate large language models (LLMs) into IE via two paradigms: direct extraction and pre-trained language model (PLM)-assisted extraction. Direct extraction has shifted from supervised fine-tuning to generative approaches, like UIE’s unified text-to-structure framework~\cite{DBLP:conf/ccks/ZhaoWK22} and InstructUIE’s multi-task instruction tuning~\cite{DBLP:journals/corr/abs-2304-08085,DBLP:conf/coling/LiaoDH025,chunkuie2025}. PLM-assisted methods combine LLMs with smaller pretrained models, either using LLMs as primary extractors with PLMs for retrieval/calibration~\cite{DBLP:journals/corr/abs-2310-18463,DBLP:conf/acl/ZhangBCSV24}, or employing PLMs for extraction with LLMs as data generators or validators~\cite{DBLP:conf/naacl/ZaratianaTHC24,DBLP:conf/acl/XuWLD0M23}. Hybrid evaluation techniques have also emerged from this collaboration~\cite{DBLP:conf/coling/FanL0YHL24}.

Recent advancements have integrated LLMs into IE through two primary strategies: direct extraction and pre-trained language model (PLM)-assisted extraction. Direct extraction has evolved from supervised fine-tuning to generative paradigms, exemplified by UIE's unified text-to-structure framework \cite{DBLP:conf/ccks/ZhaoWK22} and InstructUIE's multi-task instruction tuning \cite{DBLP:journals/corr/abs-2304-08085,DBLP:conf/coling/LiaoDH025,chunkuie2025}. PLM-assisted methods adopt hybrid architectures where LLMs either serve as primary extractors with PLMs for retrieval/calibration \cite{DBLP:journals/corr/abs-2310-18463,DBLP:conf/acl/ZhangBCSV24}, or where PLMs perform extraction while LLMs generate synthetic training data \cite{DBLP:conf/naacl/ZaratianaTHC24,DBLP:conf/acl/XuWLD0M23}. This synergy has also spurred novel hybrid evaluation protocols \cite{DBLP:conf/coling/FanL0YHL24}.

Unlike existing approaches that treat fine-tuning and in-context learning as separate paradigms, our SCIR framework uniquely unifies these mechanisms within an LLM-assisted architecture, demonstrating superior performance through rigorous experimental validation.

%In summary, LLM-based information extraction is evolving through three pivotal developments: transitioning from supervised learning to generative frameworks for unified multi-task processing, advancing collaborative architectures combining LLMs with specialized PLMs for enhanced capability synergy, and converging open-domain flexibility with schema-based standardization. Distinct from existing approaches, our SCIR uniquely integrates fine-tuning and in-context learning within a LLM-assisted paradigm, proving more effective through experimental validation.

	\section{Methodology}
	\label{sec:method}
	% !TEX spellcheck = en_US
% !TeX root = main.tex

\begin{figure*}[h]
	\centering
	\includegraphics[width=0.95\textwidth, keepaspectratio]{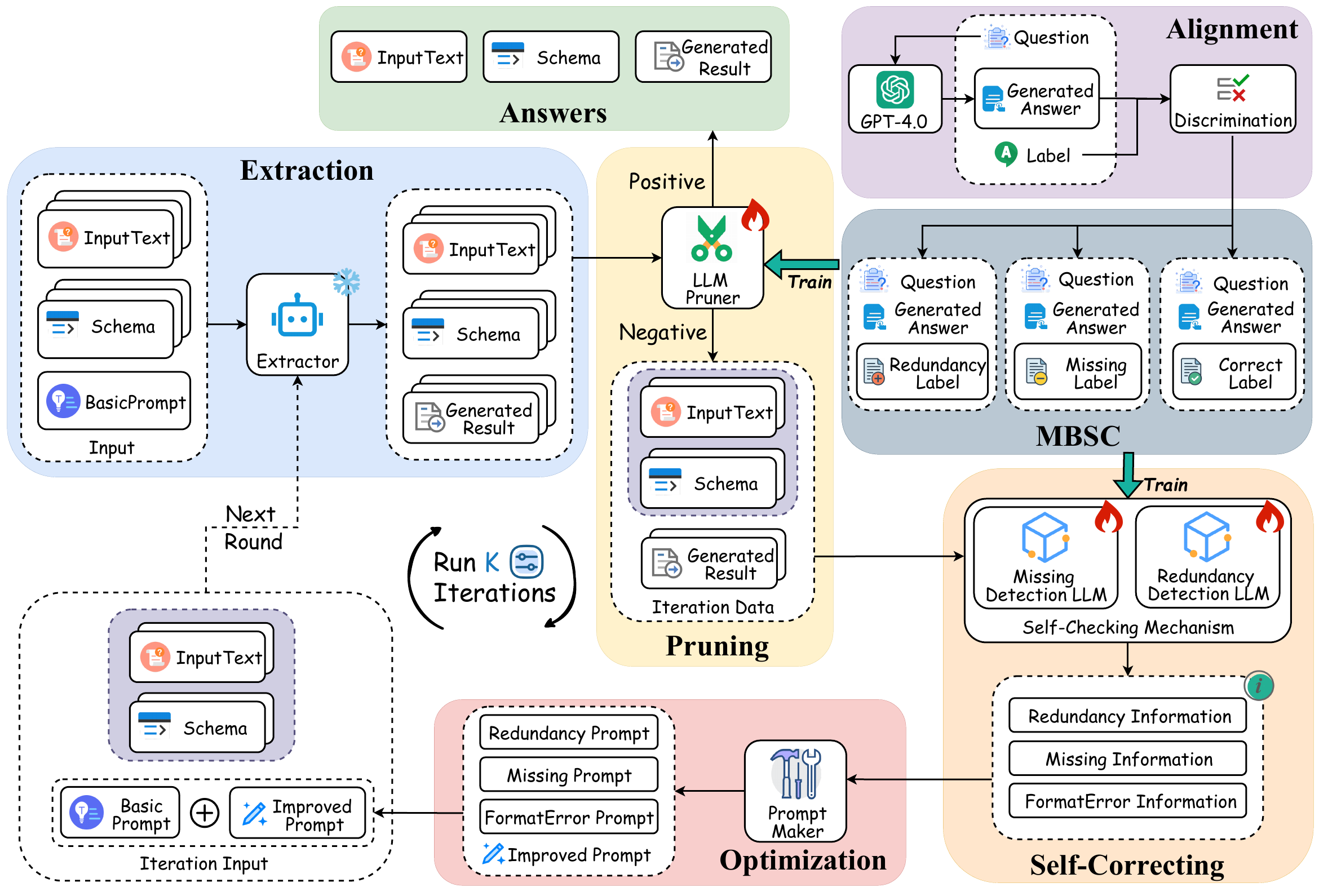}
	\caption{The architecture of SCIR, the number of iterations K is a hyperparameter.}
	\label{fig:model}
\end{figure*}

We first introduce the proposed SCIR framework, and then detail the construction of the MBSC dataset.

\subsection{SCIR Framework}
\label{sec: SCIR}

%As shown in Figure~\ref{fig:model}, SCIR improves content accuracy and completeness through iterative optimization. Its core architecture consists of four key modules: (1) information extraction module, (2) iterative pruning module, (3) dual-path self-correction module, and (4) feedback-driven optimization module. This section will sequentially detail the structure and function of each module. The corresponding algorithm pseudocode is presented in Algorithm~\ref{alg:SCIR}.

As depicted in Figure~\ref{fig:model}, the SCIR framework enhances extraction quality through iterative refinement. The architecture consists of four core components: (1) an information extraction module, (2) an iterative pruning module, (3) a dual-path self-correcting module, and (4) a feedback-driven optimization module. These components operate in a synergistic pipeline to progressively improve extraction results through coordinated interactions. Below, we provide detailed descriptions of each module's structural design and functional mechanisms.
The corresponding algorithmic implementation is detailed in Algorithm~\ref{alg:SCIR}.

\subsubsection{Information Extraction Module} %Unlike traditional approaches that rely on fine-tuning large models with vertical domain data for extraction, we adopt a more flexible extraction paradigm. As shown in the `Extraction' module of the Figure~\ref{fig:model}, our information extractor can be: an untrained base model~\cite{yang2025qwen3technicalreport,deepseekai2025deepseekr1incentivizingreasoningcapability,Meta2024Llama}, a domain-specific fine-tuned model~\cite{DBLP:journals/corr/abs-2412-20005}, or an existing IE framework~\cite{DBLP:conf/coling/LiaoDH025}. During initial extraction, the extractor performs preliminary IE on raw text and basic instruction templates; subsequent iterative extractions dynamically adjust results by incorporating optimized prompts filtered through the detection module. This design offers dual advantages: achieving plug-and-play compatibility with new models through standardized interfaces while significantly enhancing existing models' extraction performance via in-context learning capabilities.
The Information Extraction Module is shown as the \textsf{Extraction} block in Figure~\ref{fig:model}. This module introduces a flexible paradigm distinct from conventional methods requiring domain-specific fine-tuning, supporting three extractor configurations: (1) untrained LLMs ~\cite{yang2025qwen3technicalreport,deepseekai2025deepseekr1incentivizingreasoningcapability,Meta2024Llama}; (2) domain-adapted fine-tuned variants~\cite{DBLP:journals/corr/abs-2412-20005}; and (3) existing IE frameworks~\cite{DBLP:conf/coling/LiaoDH025}. Initial extraction performs preliminary information extraction on raw text using basic instruction templates, while subsequent iterations dynamically refine results through optimized prompts filtered by the detection module. This design achieves dual advantages: plug-and-play compatibility with new models via standardized interfaces, enabling rapid integration without architectural modifications, and performance enhancement of existing models through in-context learning capabilities that iteratively guide the extraction process with refined prompts.

\begin{algorithm}[t]
    \caption{Overall workflow of SCIR}
    \small
    \label{alg:SCIR}
    \renewcommand{\algorithmicrequire}{\textbf{Input:}}
    \renewcommand{\algorithmicensure}{\textbf{Output:}}
    \begin{algorithmic}[1]
        \REQUIRE $Max\:Iterations:K$, $Basic\:Prompt:B_{prompt}$, $Data$
        \ENSURE $Answer_{set}$
        \STATE $Answer_{set} = \{\}$
        \STATE $Round_{prompt} = B_{prompt}$
        \STATE $round = 0$
        \WHILE{$Data \neq \emptyset \; and \; round \le K$}
            \STATE $Gen_{result} \leftarrow LLM(Round_{prompt} \oplus Data)$
            \STATE $Check_{pos},Check_{neg} \leftarrow Prun_{model}(Gen_{result})$
            \IF {$Check_{pos} \neq \emptyset$}
                \STATE $ Answer_{set} \leftarrow Answer_{set} \cup Check_{pos}$
                \STATE $ Data \leftarrow Data - Check_{pos}$
            \ENDIF
            \IF {$Check_{neg} \neq \emptyset$}
                \STATE $Red_{set} \leftarrow Red_{model}(Check_{neg})$
                \STATE $Mis_{set}\leftarrow Mis_{model}(Check_{neg})$
                \STATE $Red_{prompt} \leftarrow RP_{maker}(Red_{set})$
                \STATE $Mis_{prompt} \leftarrow MP_{maker}(Mis_{set})$
            \ENDIF
            \STATE $Iteration_{prompt} \leftarrow Red_{prompt} \cup Mis_{prompt}$
            \STATE $Round_{prompt} \leftarrow B_{prompt} \oplus Iteration_{prompt}$
            \STATE $round \leftarrow round + 1$
        \ENDWHILE
        \IF {$round = K$}
            \STATE $Answer_{set} \leftarrow Answer_{set} \cup Check_{neg}$
        \ENDIF
        \RETURN $Answer_{set}$
    \end{algorithmic}
\end{algorithm}

\subsubsection{Result Pruning Module} %Considering that the extraction results may contain correct answers, detecting these results and using them as the final output can significantly improve extraction efficiency. Therefore, we designed a "pruning" module to filter out data that does not require iteration.  As shown in the Pruning module in Figure~\ref{fig:model}, this pruner is trained on the MBSC dataset using the Qwen3-4B ~\cite{yang2025qwen3technicalreport} model. During the extraction phase, the obtained data is divided by the pruning module into two datasets: Positive results are directly output as extraction results, while Negative results enter the dual-path self-correction module to rectify extraction errors.
The Result Pruning Module addresses the efficiency challenge in iterative extraction by strategically identifying correct results for early termination. Recognizing that raw extraction outputs may already contain valid answers, we designed a discriminative pruning mechanism to bypass unnecessary iterations for confirmed-correct data. As depicted in the \textsf{Pruning} block of Figure~\ref{fig:model}, this component employs a Qwen3-4B~\cite{yang2025qwen3technicalreport}-based classifier trained on the MBSC dataset to partition extraction results into two categories: Positive samples meeting confidence thresholds are immediately output as final results, while Negative samples with potential errors are routed to the Dual-Path Self-Correcting module for refinement.
This binary classification method effectively reduces computational load via early termination while ensuring accuracy by rectifying ambiguous results.
 %This binary classification achieves dual optimization: reducing computation by terminating valid extractions early, and ensuring accuracy by correcting ambiguous results. %The confidence threshold dynamically adjusts during iterations based on self-correction feedback, maintaining an adaptive efficiency-precision balance throughout refinement.

\subsubsection{Dual-Path Self-Correcting Module} 
The Dual-Path Self-Correcting Module enhances iterative extraction by simultaneously resolving redundancy and omission issues through a joint detection architecture. As illustrated in Figure~\ref{fig:model}'s \textsf{Self-Correcting} block, this system employs two parallel paths: (1)  \textbf{Redundancy Detection Path} systematically analyzes extraction outputs to identify and aggregate Redundancy Information into a structured Redundancy set, while (2) \textbf{Missing Detection Path} verifies logical and contextual coherence, generating a missing set. Both paths are also Qwen3-4B models fine-tuned by MBSC dataset. Additionally, any format violations detected during analysis are compiled into a FormatError set. Redundancy, Missing and FormatError sets provide multi-dimensional correction signals that enable precise error localization while maintaining full interpretability of the optimization process. The dual-path design achieves synergistic effects: the redundancy path ensures output conciseness, and the missing path guarantees completeness, collectively enhancing extraction quality through interpretable iterative refinement.

\subsubsection{Feedback-Driven Optimization} 
The Feedback-Driven Optimization module implements a closed-loop refinement system by injecting detection results into adaptive prompts. As shown in Figure~\ref{fig:model}, three diagnostic feedback streams generate specialized prompts: Redundancy Prompt, Missing Prompt, and FormatError Prompt. These prompts are dynamically fused with Basic Prompt into composite prompts, while preserving contextual semantics that guide LLM iterations, with each cycle incorporating updated feedback for progressive quality improvement.

\subsection{MBSC Dataset}
\label{sec: mbsf}
Traditional IE models rely on manually cleaned data for fine-tuning, which makes it difficult to effectively align with the model's prediction preferences. To align with the preferences of the Pruning module and the Self-Correcting module with the model's preferences, we created the MBSC dataset for training the model's detection capability and preference alignment. As shown in the \textsf{Alignment} block of Figure~\ref{fig:model}, built upon the IEPile dataset~\cite{DBLP:conf/acl/GuiYYZSLC24}, MBSC employs GPT-4 to generate predictions for each IE question and systematically identifies three discrepancy patterns by comparing to original labels: missing information, redundant content, and correct matches. We employ GPT-4 as the generative model, as it represents the state-of-the-art in LLMs. Its characteristic errors often reflect common limitations across comparable models. For missing/redundant cases, we augment the label with absent content and a missing/redundant marker. For correct matches, we apply the \emph{\textless Correct\textgreater} tag. This model behavior-driven approach constructs a training set that precisely evaluates content completeness while capturing common generation flaws even in advanced models like GPT-4. Unlike traditional synthetic methods relying on random modifications, MBSC's targeted label reconstruction mechanism achieves deeper alignment between detection capabilities and generative preferences, substantially improving the Self-Checking Mechanism's robustness through exposure to authentic error patterns from state-of-the-art models.

	\section{Experiments}
	\label{sec:experiment}
	% !TEX spellcheck = en_US
% !TeX root = main.tex

%This section describes our experiment methodology, which includes: the datasets, evaluation metrics, baselines, and experimental settings.

\setlength{\tabcolsep}{6pt}
\renewcommand{\arraystretch}{1}
 \begin{table}[t]
 	\centering
 	\small
 	\begin{tabular}{l | c | c | c }
 		\hline
 		\textbf{Dataset} & \textbf{Lang.} & \textbf{Task} & \textbf{Domain} \\ \hline
 		COAE2016   & ZH  & RE  & Web Text      \\
 		SKE2020    & ZH  & RE  & Commercial    \\
 		Wiki-ZSL   & EN  & RE  & Encyclopedia  \\
 		FewRel~\cite{DBLP:conf/emnlp/HanZYWYLS18}& EN  & RE  & Experiment\\ \hline  
 		Boson      & ZH  & NER & Financial     \\
 		Weibo      & ZH  & NER & Social Media  \\
 		CrossNER~\cite{DBLP:conf/aaai/Liu0YDJCMF21}& EN  & NER & Experiment    \\ \hline  
 		CCF Law    & ZH  & EE  & Legal         \\ 
 		FewFC~\cite{DBLP:conf/acl/ShengGYLHWLX21}& ZH  & EE  & Judicial      \\ 
 		RAMS       & EN  & EE  & News          \\ 
 		WikiEvents & EN  & EE  & Encyclopedia  \\ \hline  
 	\end{tabular}
 	\caption{Dataset details.}
 	\label{tab:dataset}
 \end{table}

\setlength{\tabcolsep}{8pt}
\renewcommand{\arraystretch}{1}
\begin{table*}[t]
	\small
	\centering
	\begin{tabular}{l|c c|c|c c|c|c c|c}
		\hline
		\multicolumn{1}{l|}{}  & \multicolumn{3}{c|}{EE} & \multicolumn{3}{c|}{NER} & \multicolumn{3}{c}{RE} \\ \hline
		Model & \begin{tabular}[c]{@{}c@{}}CCF Law\end{tabular} & \begin{tabular}[c]{@{}c@{}}FewFC\end{tabular} & \begin{tabular}[c]{@{}c@{}}Avg\end{tabular} & \begin{tabular}[c]{@{}c@{}}Weibo\end{tabular} & \begin{tabular}[c]{@{}c@{}}Boson\end{tabular} & \begin{tabular}[c]{@{}c@{}}Avg\end{tabular} & \begin{tabular}[c]{@{}c@{}}COAE2016\end{tabular} & \begin{tabular}[c]{@{}c@{}}SKE2020\end{tabular} & \begin{tabular}[c]{@{}c@{}}Avg\end{tabular} \\ \hline
		LLama3.1 & 31.57 & 32.52 & 32.05 & 17.02 & 29.74 & 23.38 & 28.66 & 34.74 & 31.70 \\ 
		Qwen3 & 34.99 & 41.29 & 38.14 & 19.01 & 35.42 & 27.21 & 25.49 & 36.73 & 31.11 \\ 
		DeepSeek-R1 & 38.79 & 40.81 & 39.80 & 24.66 & 40.67 & 32.67 & 32.69 & 45.32 & 39.00 \\ \hline
		YAYI-UIE & 12.87 & \underline{81.28} & 47.08 & 36.46 & 49.25 & 42.86 & 19.97 & 70.8 & 45.39 \\
		IEPile-LLama2 & 59.90 & 70.10 & 65.00 & 34.97 & 54.45 & 44.71 & 46.70 & 72.18 & 59.44 \\
		ChunkUIE & 61.41 & 79.75 & 70.58 & 35.11 & 59.00 & 47.06 & 48.20 & 70.91 & 59.56 \\
		OneKE & 62.19 & 80.11 & 71.15 & 35.06 & \textbf{72.61} & 53.84 & 49.83 & 72.61 & 61.22 \\  \hline
		SCIR-LLama3.1 & 60.99 & 63.75 & 62.37 & 31.21 & 58.03 & 44.62 & \textbf{54.26} & 67.68 & 60.97 \\
		SCIR-Qwen3 & \underline{65.86} & 77.42 & \underline{71.64} & 34.11 & 66.08 & 50.10 & 46.96 & 66.51 & 56.74 \\
		SCIR-DeepSeek-R1 & 62.45 & 68.20 & 65.32 & \underline{39.68} & 68.45 & \underline{54.07} & 52.45 & \textbf{74.48} & \textbf{63.47} \\ \hline
		SCIR-OneKE & \textbf{67.01} & \textbf{85.10} & \textbf{76.05} & \textbf{41.35} & \underline{68.58} & \textbf{54.97} & \underline{51.05} & \underline{73.81} & \underline{62.43} \\ \hline
	\end{tabular}
	\caption{Performance comparison in Chinese IE Tasks. Best results are in bold and the second best are underlined.}
	\label{tab:zh}
\end{table*}

\setlength{\tabcolsep}{12pt}
\renewcommand{\arraystretch}{1}
\begin{table*}[t]
	\small
	\centering
	\begin{tabular}{l|c c|c|c|c c|c}
		\hline
		\multicolumn{1}{l|}{}  & \multicolumn{3}{c|}{EE} & \multicolumn{1}{c|}{NER} & \multicolumn{3}{c}{RE} \\ \hline
		Model & \begin{tabular}[c]{@{}c@{}}RAMS\end{tabular} & \begin{tabular}[c]{@{}c@{}}WikiEvents\end{tabular} & \begin{tabular}[c]{@{}c@{}}Avg\end{tabular} & \begin{tabular}[c]{@{}c@{}}CrossNER\end{tabular} & \begin{tabular}[c]{@{}c@{}}Wiki-ZSL\end{tabular} & \begin{tabular}[c]{@{}c@{}}FewRel\end{tabular} & \begin{tabular}[c]{@{}c@{}}Avg\end{tabular} \\ \hline
		LLama3.1 & 10.26 & 7.13 & 8.69 & 26.65 & 13.65 & 19.14 & 16.39 \\ 
		Qwen3 & 12.67 & 8.27 & 10.47 & 30.90 & 16.01 & 17.75 & 16.88 \\ 
		DeepSeek-R1 & 13.02 & 8.68 & 10.85 & 37.84 & 24.98 & 21.79 & 23.38 \\ \hline
		YAYI-UIE & 18.87 & 10.97 & 14.92 & 50.39 & 41.07 & 36.09 & 38.58 \\
		IEPile-LLama2 & 23.62 & 13.93 & 18.78 & 56.50 & 36.18 & 37.14 & 36.66 \\
		ChunkUIE & 19.71 & 8.67 & 14.19 & 58.13 & 32.23 & 35.76 & 33.99 \\ 
		OneKE & 22.58 & 12.43 & 17.51 & 60.91 & 42.18 & 39.19  & 40.69  \\ 
		RUIE & 26.06 & \underline{40.64} & \underline{33.35} & \underline{65.41} & \underline{53.16} & \underline{49.93} & \underline{51.55} \\ \hline
		SCIR-LLama3.1 & 20.53 & 13.68 & 17.11 & 54.71 & 28.01 & 38.60 & 33.31 \\
		SCIR-Qwen3 & 24.70 & 15.61 & 20.15 & 61.51 & 31.64 & 34.67 & 33.17 \\
		SCIR-DeepSeek-R1 & 21.21 & 13.54 & 17.37 & 62.67 & 42.34 & 36.55 & 39.43 \\ \hline
		SCIR-OneKE & \textbf{27.04} & 16.97 & 22.00 & 63.70 & 43.41 & 45.16 & 44.29 \\
		SCIR-RUIE & \underline{26.94} & \textbf{45.74} & \textbf{36.34} & \textbf{65.54} & \textbf{53.71} & \textbf{55.02} & \textbf{54.37} \\ \hline
	\end{tabular}
	\caption{Performance comparison in English IE Tasks. Best results are in bold and the second best are underlined.}
	\label{tab:en}
\end{table*}

\subsection{Experimental Setup}

\subsubsection{Datasets \& Metrics}  

To comprehensively validate our approach, we selected 11 benchmark datasets spanning three core IE tasks: (1) \textbf{Event Extraction}: CCF Law, FewFC, RAMS, and WikiEvents; (2) \textbf{Named Entity Recognition}: Boson, Weibo, and CrossNER; and (3) \textbf{Relation Extraction}: COAE2016, SKE2020, and FewRel. Dataset details are summarized in Table~\ref{tab:dataset}. Crucially, all experiments adhered to a strict zero-shot evaluation protocol where test sets were entirely excluded from training data. For quantitative assessment, we adopted the span-based Micro-F1 metric as the primary evaluation criterion. Based on the iterative round experiment results, we set the number of iterations to 2, achieving optimal efficiency without compromising effectiveness, and all results are based on this iteration round.

\subsubsection{Baselines} We compare SCIR with several representative baselines, including untuned LLMs (LLama3.1, Qwen3, DeepSeek-R1),  domain-specific models (YAYI-UIE, IEPile-LLama2, ChunkUIE and OneKE) and current mainstream IE frameworks (RUIE):

\begin{itemize}
\item \textbf{LLama3.1-8B}~\cite{Meta2024Llama}\footnote{https://github.com/meta-llama/llama3.}: An open-source multilingual  LLM released by Meta, which is suitable for multilingual conversations and text generation tasks.

\item \textbf{Qwen3-8B}~\cite{yang2025qwen3technicalreport}\footnote{https://github.com/QwenLM/Qwen3.}: Alibaba's 8-billion parameter dense model featuring RL-optimized performance in STEM and coding domains.

\item \textbf{DeepSeek-R1-Distill-Qwen3-8B}~\cite{deepseekai2025deepseekr1incentivizingreasoningcapability} \footnote{https://github.com/deepseek-ai/DeepSeek-R1.}: This model is developed by DeepSeek through Chain-of-Thought distillation from its flagship model DeepSeek-R1-0528.

\textbf{YAYI-UIE}~\cite{DBLP:journals/corr/abs-2312-15548}\footnote{https://github.com/wenge-research/YAYI-UIE}: A unified IE system trained on over 1 million human-annotated samples, supporting structured extraction across 12 distinct domains.

\item \textbf{IEPile-LLama2}~\cite{DBLP:conf/acl/GuiYYZSLC24}\footnote{https://github.com/zjunlp/IEPile.}: A LLaMA2-13B model fine-tuned using LoRA on the IEPile dataset, demonstrating robust bilingual IE capabilities.

\item \textbf{ChunkUIE}~\cite{chunkuie2025}\footnote{https://github.com/ChunkUIE/chunkuie.}: Implements chunked instruction processing and hard negative sampling to address semantic ambiguity in bilingual IE tasks.

\item \textbf{OneKE}~\cite{DBLP:journals/corr/abs-2412-20005}\footnote{https://github.com/zjunlp/OneKE.}: A large-scale model IE framework featuring bilingual support and generalization capabilities for multiple domains and tasks.

\item \textbf{RUIE}~\cite{DBLP:conf/coling/LiaoDH025}\footnote{https://github.com/OStars/RUIE.}: An English-only extraction framework utilizing BM25 sparse retrieval for efficient candidate screening.
\end{itemize}
Moreover, to assess SCIR's  plug-and-play capability, we implemented two sets of its variants for comparison: 1) SCIR based on untuned LLMs (SCIR-LLama3.1, SCIR-Qwen3 and SCIR-DeepSeek-R1), and 2) SCIR integrated with representative domain-specific models (SCIR-OneKE) and mainstream IE frameworks (SCIR-RUIE).

\subsection{Main Results}

\subsubsection{Chinese Dataset Performance}
As presented in Table~\ref{tab:zh}, the SCIR framework demonstrates outstanding performance on Chinese datasets. In EE tasks, SCIR achieves exceptional results when integrated with vertical domain-specific models, while showing slightly reduced performance when combined with base LLMs—likely due to prompt comprehension challenges caused by event structure complexity. Nevertheless, both integration approaches significantly outperform baseline systems. For NER tasks, SCIR's performance improvement appears relatively modest, potentially constrained by the inherent simplicity of the task. In RE tasks, SCIR excels particularly well, substantially enhancing the typically mediocre performance of LLMs in this task. We attribute this improvement to the Dual-Path Self-Correcting mechanism's precise control over redundant and missing content, effectively unleashing the potential of LLMs. Notably, SCIR consistently delivers significant performance gains across different extractors, demonstrating its excellent plug-and-play capability.

%As demonstrated in Table~\ref{tab:zh}, the SCIR framework exhibits remarkable efficacy across Chinese datasets, with distinct performance patterns emerging across different tasks. In EE tasks, SCIR achieves superior results when integrated with domain-specific models, though its performance with base LLMs shows a slight decline—likely attributable to the challenges in prompt interpretation arising from complex event structures. Nevertheless, both integration strategies substantially outperform baseline systems, underscoring SCIR's robust adaptability. For NER, the observed performance gains are relatively moderate, possibly due to the inherent simplicity of the task limiting the framework's optimization potential. In contrast, SCIR demonstrates exceptional proficiency in RE tasks, significantly elevating the typically subpar performance of LLMs in this domain through its Dual-Path Self-Correcting Mechanism, which effectively regulates redundant and missing content to unlock the full potential of LLMs. Notably, SCIR consistently delivers substantial performance improvements across diverse extractors, providing compelling evidence of its exceptional plug-and-play capability.

\setlength{\tabcolsep}{4pt}
\renewcommand{\arraystretch}{1}
\begin{table}[t]
	\centering
	\footnotesize
	\begin{tabular}{c | l | c  c  c  c}
		\hline
		\textbf{Task} & \textbf{Dataset} & \textbf{w/o Both} & \textbf{w/o Red} & \textbf{w/o Mis} & \textbf{FULL} \\ \hline
		\multirow{4}{*}{EE}   
		& CCF Law    & 34.99 & 63.62 & 62.73 & \textbf{65.86} \\
		& FewFC      & 41.29 & 70.12 & 74.44 & \textbf{77.42} \\
		& RAMS       & 12.67 & 21.79 & 22.71 & \textbf{24.70} \\
		& WikiEvents & 8.27 & 11.14 & 13.32 & \textbf{15.61} \\ \hline 
		\multirow{3}{*}{NER}  
		& boson      & 35.42 & 64.45 & 61.27 & \textbf{66.08} \\
		& WEIBONER   & 19.01 & 31.95 & 31.79 & \textbf{34.11} \\
		& CrossNER   & 30.90 & 57.05 & 58.39 & \textbf{61.51} \\ \hline
		\multirow{4}{*}{RE}   
		& COAE2016   & 25.49 & 38.55 & 42.40 & \textbf{46.96} \\
		& SKE2020    & 36.73 & 59.52 & 62.23 & \textbf{66.51} \\ 
		& Wiki-ZSL   & 16.01 & 28.45 & 27.48 & \textbf{31.64} \\ 
		& FewRel     & 17.75 & 33.11 & 28.77 & \textbf{34.67} \\ \hline  
	\end{tabular}
	\caption{Module ablation study results, where ‘w/o Red’ denotes using only the missing detection module and ‘w/o Mis’ denotes using only the redundant detection module. }
	\label{tab:comparative}
\end{table}

\subsubsection{English Dataset Performance}
As shown in Table~\ref{tab:en}, SCIR maintains excellent performance on English datasets. Benefiting from the advantages of the robust RUIE framework, SCIR achieves or approaches state-of-the-art performance across all three IE tasks. However, compared to its performance on Chinese datasets, SCIR demonstrates slightly diminished performance gains on English datasets. We attribute this primarily to the fact that the Dual-Path Self-Correcting mechanism was trained using the Qwen3-4B model, which inherently possesses stronger capabilities for Chinese tasks due to its corpus composition. Notably, SCIR successfully achieves effective integration with base LLMs,  domain-specific models, and IE frameworks, fully demonstrating its superior model generalization capability.

%In sum, the SCIR framework demonstrates significant performance advantages across both Chinese and English datasets, achieving a 5.27\% performance improvement compared to the current state-of-the-art bilingual extraction model (OneKE). In Chinese scenarios, the framework effectively enhances the capabilities of base LLMs (e.g., Qwen3, LLama3.1, DeepSeek) for complex IE tasks through its Dual-Path Self-Correcting mechanism, while maintaining excellent compatibility with vertical domain-specific models (e.g., OneKE). Meanwhile, for English tasks, SCIR successfully integrates with existing frameworks (e.g., RUIE) to achieve state-of-the-art performance. These cross-lingual and cross-task validations conclusively prove SCIR's exceptional compatibility and flexibility, enabling plug-and-play functionality with arbitrary models.

In sum, the SCIR framework exhibits substantial performance gains across Chinese and English datasets, delivering a 5.27\% improvement over current state-of-the-art bilingual extraction models (OneKE). For Chinese tasks, SCIR's Dual-Path Self-Correcting mechanism effectively augments base LLMs (Qwen3, LLaMA-3.1, DeepSeek) on complex tasks while maintaining seamless integration with domain-specific models like OneKE. For English tasks, SCIR successfully combines with existing frameworks (e.g., RUIE) to achieve cutting-edge performance. These comprehensive cross-lingual evaluations demonstrate SCIR's remarkable flexibility and compatibility, enabling true plug-and-play functionality with diverse model architectures.

\begin{figure}[t]
	\centering
	\includegraphics[width=0.95\linewidth]{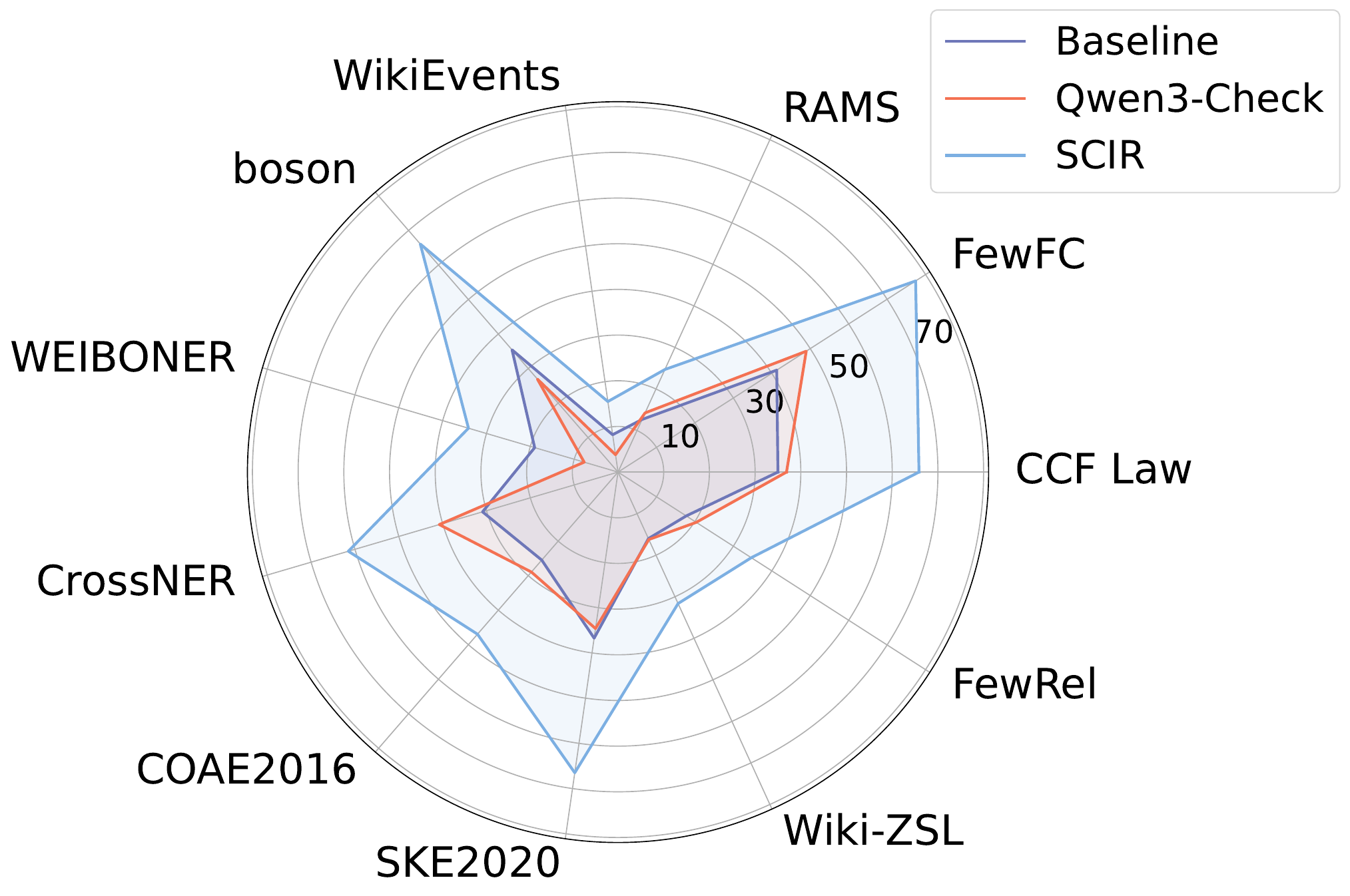}
	\caption{The figure shows the F1 scores of the three model ablation experiments on each dataset. The term "Baseline" refers to only use the LLMs to extraction, while "Qwen3-Check" denotes using an untuned Qwen3-4B model in Dual-Path Self-Correcting Module.}
	\label{fig:qwen}
\end{figure}

\subsection{Ablation Study}

\subsubsection{Module Ablation Study}
%Since both the 'Information Extraction' and 'Feedback-Driven Optimization' modules are essential components in SCIR, and the 'Result Pruning' module does not affect SCIR's performance, we conducted ablation experiments solely on the 'Dual-Path Self-Correcting Module'. Our ablation study systematically evaluates the 'Self-Checking Mechanism' through comparative analysis of three configurations: redundant detection only, missing detection only, and the complete dual-path implementation. The results in Table~\ref{tab:comparative} reveal that the full dual-path approach delivers statistically significant improvements across all tasks, with distinct patterns emerging across different domains. In EE, redundant detection proves more impactful, while missing detection shows stronger effects in NER. For RE, we observe language-dependent variations - redundant detection excels on Chinese datasets whereas missing detection performs better on English corpora. These findings conclusively validate the dual-path design, demonstrating how the two modules operate in complementary fashion to enhance overall model performance.

In this set of experiments, we focus on the `Dual-Path Self-Correcting Module' since other components are essential or performance-neutral. Our experiments compare three configurations: redundant-only, missing-only, and full dual-path detection. Table~\ref{tab:comparative} shows the complete implementation achieves statistically significant gains across all tasks, with distinct patterns emerging across different domains. In EE, redundant detection proves more impactful, while missing detection shows stronger effects in NER. For RE, we observe language-dependent variations - redundant detection excels on Chinese datasets whereas missing detection performs better on English corpora. These findings conclusively validate the dual-path design, demonstrating how the two modules operate in complementary fashion to enhance overall model performance.

% \begin{figure}[t]
% 	\centering
% 	\includegraphics[width=\linewidth]{img/5.png}
% 	\caption{Comparison between traditional Information Extraction methods and our SCIR. The figure shows the performance of each module ablation as a percentage relative to SICR as the baseline.}
% 	\label{fig:comparative}
% 	\vspace{-1em}
% \end{figure}

% \setlength{\tabcolsep}{9pt}
% \begin{table}[!b]
% 	\centering
% 	\small
% 	\begin{tabular}{l | l | c  c  c}
% 		\hline
% 		\textbf{Task} & \textbf{Dataset} & \textbf{Base} & \textbf{Qwen3} & \textbf{SCIR}\\ \hline
% 		\multirow{4}{*}{EE}   
% 		& CCF Law    & 34.99 & 36.87 & \textbf{65.86} \\
% 		& FewFC      & 41.29 & 48.96 & \textbf{77.42} \\
% 		& RAMS       & 12.67 & 14.25 & \textbf{24.70} \\
% 		& WikiEvents & 8.27  & 3.84  & \textbf{15.61} \\ \hline 
% 		\multirow{3}{*}{NER}  
% 		& boson      & 35.42 & 26.94 & \textbf{66.08} \\
% 		& WEIBONER   & 19.01 & 7.72  & \textbf{34.11} \\
% 		& CrossNER   & 30.90 & 40.77 & \textbf{61.51} \\ \hline
% 		\multirow{4}{*}{RE}   
% 		& COAE2016   & 25.49 & 28.95 & \textbf{46.96} \\
% 		& SKE2020    & 36.73 & 34.61 & \textbf{66.51} \\ 
% 		& Wiki-ZSL   & 16.01 & 16.23 & \textbf{31.64} \\ 
% 		& FewRel     & 17.75 & 20.35 & \textbf{34.67} \\ \hline  
% 	\end{tabular}
% 	\caption{Model ablation Study results. }
% 	\label{tab:qwen}
% \end{table}

\begin{figure*}[t]
	\centering
	\includegraphics[width=0.95\textwidth, keepaspectratio]{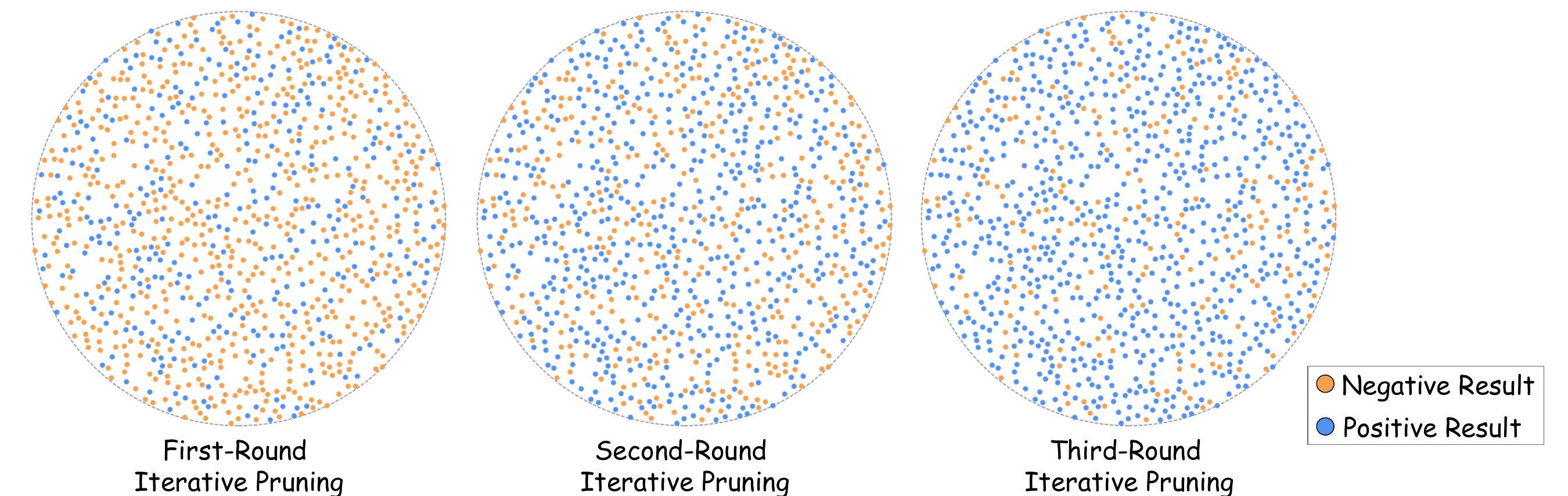}
	\caption{Iterative Experimental Results.}
	\label{fig:Pruncing}
\end{figure*}

\subsubsection{Model Ablation Study}

The model ablation study compares the untrained Qwen3-4B model as the experimental group with the same model trained on the MBSC dataset to demonstrate that SCIR's performance improvement stems from the Dual-Path Self-Correcting mechanism rather than the base model itself. The results in Figure~\ref{fig:qwen} show that the performance gains from the untrained Qwen3-4B model are negligible, while the same model trained on the MBSC dataset significantly enhances IE performance. Notably, when using the untrained Qwen3-4B model, performance degradation is observed across the WikiEvent, boson, and SKE2020 datasets, indicating that model errors are amplified iteratively due to the lack of timely correction. The ablation study confirms that SCIR's performance improvement originates from the Dual-Path Self-Correcting mechanism rather than the base model itself, while also highlighting the importance of the MBSC dataset.

% In summary, the ablation experiments demonstrate that both redundancy detection and missing detection in the Dual-Path Self-Correcting mechanism of the SCIR framework contribute significantly to performance improvement. More importantly, the superior performance of the detection model does not stem from the inherent capabilities of the Qwen3 model itself, but rather from the distillation mechanism based on the MBSC dataset, which indirectly transfers the core capabilities of GPT-4 to the detection model. This distillation strategy enables the detection model to precisely align with the error preferences of the information extraction model, thereby significantly enhancing the detection model's accuracy and robustness.

\setlength{\tabcolsep}{5pt}
\begin{table}[t]
	\centering
	\small
	\begin{tabular}{l | c c}
		\hline
		\textbf{Task} & \begin{tabular}[c]{@{}c@{}}\textbf{Time}\end{tabular} & \begin{tabular}[c]{@{}c@{}}\textbf{Performance}\end{tabular} \\ \hline
	    Event Extraction (EE)    & +13.72\% & +48.21\% \\
		Named Entity Recognition (NER)   & +11.96\% & +42.66\% \\
	    Relation Extraction (RE)   & +14.32\% & +43.68\% \\ \hline
		Average   & +13.33\% & +44.86\% \\ \hline
	\end{tabular}
	\caption{The table presents the percentage of generation time occupied by SCIR framework's iterative detection and the performance improvement achieved using the SCIR. }
	\label{tab:time}
\end{table}

\subsection{Validation of Pruning Efficacy}
%The pruning experiment validated the effectiveness of the pruner by statistically analyzing the iterative pruning performance of the SCIR-Deepseek-R1 model on the SKE2020 dataset (results shown in Figure~\ref{fig:Pruncing}). Each point in the figure represents a piece of data to be extracted, with orange dots indicating instances that were either incorrectly retained or correctly pruned, and blue dots denoting correctly pruned instances. The observation reveals a significant increase in correctly pruned points as iterations progress, demonstrating both the pruner's efficacy and the generation of more accurate results for subsequent pruning during the iterative process. These results robustly confirm the necessity of the pruner and verify the absence of error propagation or amplification throughout the iterations.
Through statistical analysis of the iterative pruning performance of SCIR-Deepseek-R1 on the SKE2020 dataset, as shown in Figure~\ref{fig:Pruncing}, we validate the effectiveness of the pruning module. Each point in the figure represents a data item to be extracted, with orange dots indicating instances that were either incorrectly retained or correctly pruned, and blue dots denoting correctly pruned instances. Notably, as iterations progress, the number of correctly pruned points increases significantly. This observation demonstrates both the pruner's practical efficacy and its ability to generate more precise pruning decisions for subsequent iterations based on prior results. These findings robustly confirm the necessity of the pruner and verify the absence of error propagation or amplification throughout the iterative process.

\subsection{Experiment on Time Costs}
%We have recorded the time costs for both training and inference. (1) For training, the SCIR method achieves convergence in 3 hours using 4 RTX4090 GPUs. In contrast, under the same hardware, traditional methods for training vertical-domain models require 22 hours, representing a time cost reduction of approximately 87\%. (2) For inference, the results are shown in Table~\ref{tab:time}, we compare the average time consumption and average performance gains of the SCIR based on DeepSeek-R1, LLama3.1, Qwen3, OneKE, and RUIE across three bilingual tasks (EE, NER and RE). The results demonstrate that by leveraging efficient pruning techniques and rapid inference capability of lightweight detectors, the SCIR framework achieves breakthrough performance gains with only minimal additional time cost overhead.
We have meticulously recorded the time costs incurred during both training and inference. (1) In terms of training,  SCIR attains convergence within 3 hours when utilizing 4 RTX4090 GPUs. By contrast, under the same hardware, traditional methods for training vertical-domain models necessitate 22 hours, translating to an approximate 87\% reduction in time cost. (2) Regarding inference, as presented in Table~\ref{tab:time}, we compare the average time consumption and the corresponding average performance enhancements of  SCIR when integrated with DeepSeek-R1, LLama3.1, Qwen3, OneKE, and RUIE across three tasks. The results reveal that, by harnessing efficient pruning techniques and the swift inference capabilities of lightweight detectors,  SCIR achieves remarkable performance improvements while introducing only a marginal increase in time cost overhead.

\begin{figure}[t]
	\raggedright
	\includegraphics[width=0.95\linewidth]{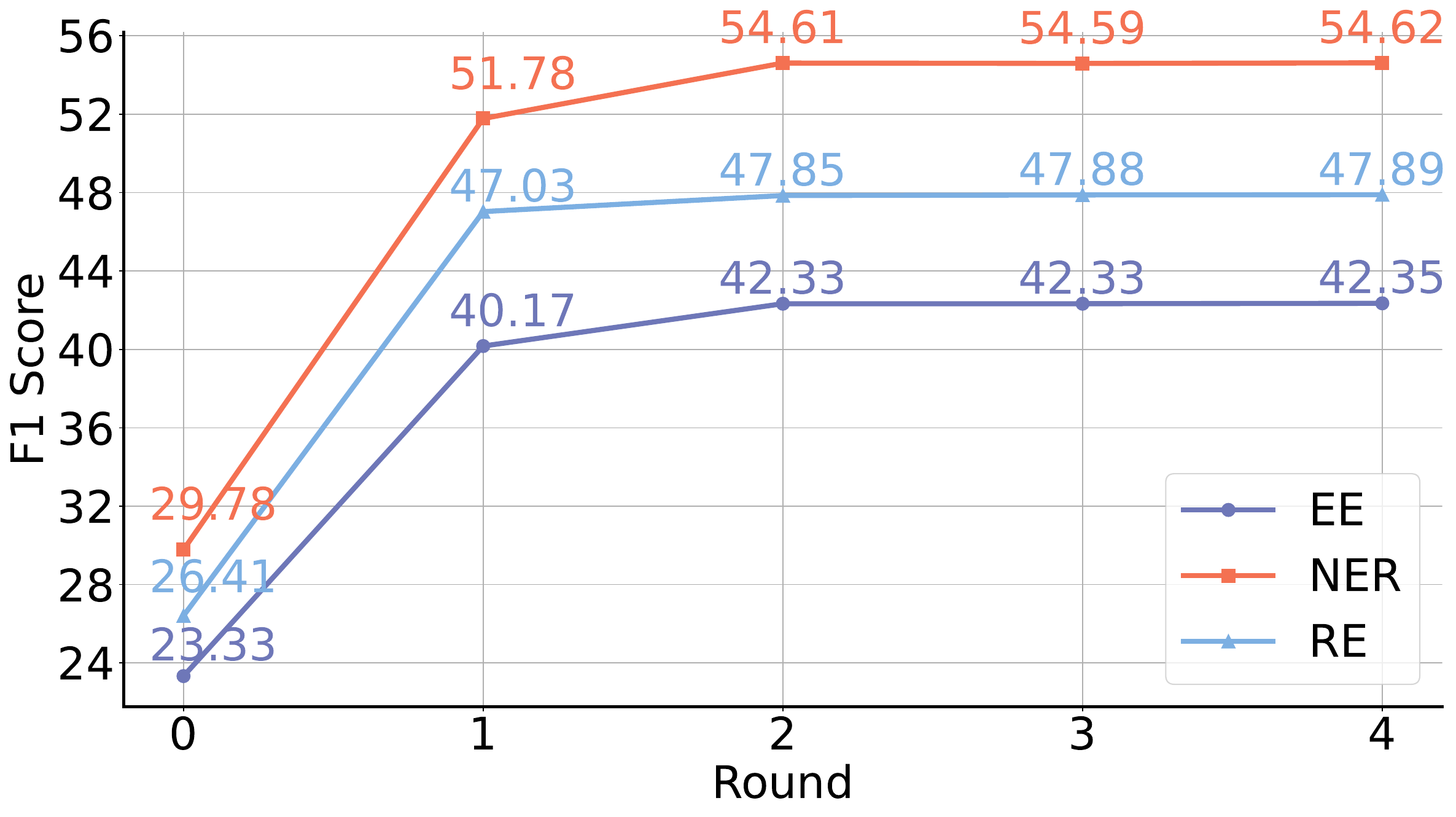}  
	\caption{Average F1 scores of the SCIR framework across multiple iterations for three IE tasks.}
	\label{fig:round}
\end{figure}

\subsection{Experiment on Iterative Round}
%We statistically analyzed the average F1 scores of the SCIR framework across multiple iterations for three IE tasks, with results shown in Figure~\ref{fig:round}. Our findings demonstrate that the SCIR framework exhibits significant performance leaps during the first two iterations, followed by diminishing returns in subsequent rounds. Accordingly, we set the iteration upper limit to two rounds for all experiments. Notably, the SCIR framework achieves breakthrough performance improvements after just a single iteration, strongly validating the framework's optimization efficacy.
We statistically analyzed the average F1 scores of  SCIR across multiple iterations for three IE tasks, as depicted in Figure~\ref{fig:round}. The results indicate that SCIR achieves significant performance gains in the first two iterations, with diminishing improvements thereafter. Consequently, we set the iteration limit to 2 for all experiments. Notably, even a single iteration yields substantial performance enhancements, strongly validating SCIR's optimization effectiveness.

	\section{Conclusion}
	\label{sec:conclusion}
	% !TEX spellcheck = en_US
% !TeX root = main.tex

%This paper proposes a Self-Correcting Iterative Refinemen (SCIR) framework for improving information extraction capabilities, and validates its performance on 11 multi-task bilingual datasets. The framework features excellent performance, low computational requirements, and easy portability, providing new design insights for lightweight and reusable information extraction frameworks.

This study proposes the Self-Correcting Iterative Refinement (SCIR) framework, whose effectiveness in enhancing IE performance has been comprehensively validated across 11 bilingual datasets covering diverse tasks. The framework demonstrates three core advantages: superior extraction accuracy, effective model preference alignment, and low-cost model portability. These characteristics position SCIR as an innovative solution for developing lightweight and reusable IE systems, while providing a new reference paradigm for future research in the field of information extraction.
	
	\bibliography{aaai2026}

@article{DBLP:journals/corr/abs-2312-00752,
  author       = {Albert Gu and
                  Tri Dao},
  title        = {Mamba: Linear-Time Sequence Modeling with Selective State Spaces},
  journal      = {CoRR},
  volume       = {abs/2312.00752},
  year         = {2023}
}

@inproceedings{DBLP:conf/aaai/JiangCXLG25,
  author       = {Chunyang Jiang and
                  Chi{-}Min Chan and
                  Wei Xue and
                  Qifeng Liu and
                  Yike Guo},
  editor       = {Toby Walsh and
                  Julie Shah and
                  Zico Kolter},
  title        = {Importance Weighting Can Help Large Language Models Self-Improve},
  booktitle    = {AAAI-25, Sponsored by the Association for the Advancement of Artificial
                  Intelligence, February 25 - March 4, 2025, Philadelphia, PA, {USA}},
  pages        = {24257--24265},
  year         = {2025}
}

@inproceedings{DBLP:conf/iclr/0001Z0S23,
  author       = {Zhuosheng Zhang and
                  Aston Zhang and
                  Mu Li and
                  Alex Smola},
  title        = {Automatic Chain of Thought Prompting in Large Language Models},
  booktitle    = {The Eleventh International Conference on Learning Representations,
                  {ICLR} 2023, Kigali, Rwanda, May 1-5, 2023},
  publisher    = {OpenReview.net},
  year         = {2023}
}

@inproceedings{DBLP:conf/coling/Xiong0XC25,
  author       = {Tianheng Xiong and
                  Wei Wei and
                  Kaihe Xu and
                  Dangyang Chen},
  editor       = {Owen Rambow and
                  Leo Wanner and
                  Marianna Apidianaki and
                  Hend Al{-}Khalifa and
                  Barbara Di Eugenio and
                  Steven Schockaert},
  title        = {{SA-DETR:} Span Aware Detection Transformer for Moment Retrieval},
  booktitle    = {Proceedings of the 31st International Conference on Computational
                  Linguistics, {COLING} 2025, Abu Dhabi, UAE, January 19-24, 2025},
  pages        = {7634--7647},
  year         = {2025}
}

@inproceedings{DBLP:conf/iclr/WanH0QB024,
  author       = {Fanqi Wan and
                  Xinting Huang and
                  Deng Cai and
                  Xiaojun Quan and
                  Wei Bi and
                  Shuming Shi},
  title        = {Knowledge Fusion of Large Language Models},
  booktitle    = {The Twelfth International Conference on Learning Representations,
                  {ICLR} 2024, Vienna, Austria, May 7-11, 2024},
  publisher    = {OpenReview.net},
  year         = {2024}
}

@article{DBLP:journals/corr/abs-2309-13064,
  author       = {Yi Yang and
                  Yixuan Tang and
                  Kar Yan Tam},
  title        = {InvestLM: {A} Large Language Model for Investment using Financial
                  Domain Instruction Tuning},
  journal      = {CoRR},
  volume       = {abs/2309.13064},
  year         = {2023}
}

@article{DBLP:journals/fcsc/MaoGFXMHG25,
  author       = {Yuren Mao and
                  Yuhang Ge and
                  Yijiang Fan and
                  Wenyi Xu and
                  Yu Mi and
                  Zhonghao Hu and
                  Yunjun Gao},
  title        = {A survey on LoRA of large language models},
  journal      = {Frontiers Comput. Sci.},
  volume       = {19},
  number       = {7},
  pages        = {197605},
  year         = {2025}
}

@inproceedings{DBLP:conf/emnlp/Qi0W00L24,
  author       = {Yunjia Qi and
                  Hao Peng and
                  Xiaozhi Wang and
                  Bin Xu and
                  Lei Hou and
                  Juanzi Li},
  editor       = {Yaser Al{-}Onaizan and
                  Mohit Bansal and
                  Yun{-}Nung Chen},
  title        = {{ADELIE:} Aligning Large Language Models on Information Extraction},
  booktitle    = {Proceedings of the 2024 Conference on Empirical Methods in Natural
                  Language Processing, {EMNLP} 2024, Miami, FL, USA, November 12-16,
                  2024},
  pages        = {7371--7387},
  year         = {2024}
}

@inproceedings{DBLP:conf/aaai/Lou0DJLH0023,
  author       = {Jie Lou and
                  Yaojie Lu and
                  Dai Dai and
                  Wei Jia and
                  Hongyu Lin and
                  Xianpei Han and
                  Le Sun and
                  Hua Wu},
  editor       = {Brian Williams and
                  Yiling Chen and
                  Jennifer Neville},
  title        = {Universal Information Extraction as Unified Semantic Matching},
  booktitle    = {Thirty-Seventh {AAAI} Conference on Artificial Intelligence, {AAAI}
                  2023, Thirty-Fifth Conference on Innovative Applications of Artificial
                  Intelligence, {IAAI} 2023, Thirteenth Symposium on Educational Advances
                  in Artificial Intelligence, {EAAI} 2023, Washington, DC, USA, February
                  7-14, 2023},
  pages        = {13318--13326},
  year         = {2023}
}

@inproceedings{DBLP:conf/emnlp/Ma0HS23,
  author       = {Yubo Ma and
                  Yixin Cao and
                  Yong Hong and
                  Aixin Sun},
  editor       = {Houda Bouamor and
                  Juan Pino and
                  Kalika Bali},
  title        = {Large Language Model Is Not a Good Few-shot Information Extractor,
                  but a Good Reranker for Hard Samples!},
  booktitle    = {Findings of the Association for Computational Linguistics: {EMNLP}
                  2023, Singapore, December 6-10, 2023},
  pages        = {10572--10601},
  year         = {2023}
}

@inproceedings{DBLP:conf/acl/0001LDXLHSW22,
  author       = {Yaojie Lu and
                  Qing Liu and
                  Dai Dai and
                  Xinyan Xiao and
                  Hongyu Lin and
                  Xianpei Han and
                  Le Sun and
                  Hua Wu},
  editor       = {Smaranda Muresan and
                  Preslav Nakov and
                  Aline Villavicencio},
  title        = {Unified Structure Generation for Universal Information Extraction},
  booktitle    = {Proceedings of the 60th Annual Meeting of the Association for Computational
                  Linguistics (Volume 1: Long Papers), {ACL} 2022, Dublin, Ireland,
                  May 22-27, 2022},
  pages        = {5755--5772},
  year         = {2022}
}

@inproceedings{DBLP:conf/naacl/TedeschiN22,
  author       = {Simone Tedeschi and
                  Roberto Navigli},
  editor       = {Marine Carpuat and
                  Marie{-}Catherine de Marneffe and
                  Iv{\'{a}}n Vladimir Meza Ru{\'{\i}}z},
  title        = {MultiNERD: {A} Multilingual, Multi-Genre and Fine-Grained Dataset
                  for Named Entity Recognition (and Disambiguation)},
  booktitle    = {Findings of the Association for Computational Linguistics: {NAACL}
                  2022, Seattle, WA, United States, July 10-15, 2022},
  pages        = {801--812},
  year         = {2022}
}

@inproceedings{DBLP:conf/acl/ZhengWBHZX17,
  author       = {Suncong Zheng and
                  Feng Wang and
                  Hongyun Bao and
                  Yuexing Hao and
                  Peng Zhou and
                  Bo Xu},
  editor       = {Regina Barzilay and
                  Min{-}Yen Kan},
  title        = {Joint Extraction of Entities and Relations Based on a Novel Tagging
                  Scheme},
  booktitle    = {Proceedings of the 55th Annual Meeting of the Association for Computational
                  Linguistics, {ACL} 2017, Vancouver, Canada, July 30 - August 4, Volume
                  1: Long Papers},
  pages        = {1227--1236},
  year         = {2017}
}

@inproceedings{DBLP:conf/naacl/WangSLOWZLWG25,
  author       = {Shuhe Wang and
                  Xiaofei Sun and
                  Xiaoya Li and
                  Rongbin Ouyang and
                  Fei Wu and
                  Tianwei Zhang and
                  Jiwei Li and
                  Guoyin Wang and
                  Chen Guo},
  editor       = {Luis Chiruzzo and
                  Alan Ritter and
                  Lu Wang},
  title        = {{GPT-NER:} Named Entity Recognition via Large Language Models},
  booktitle    = {Findings of the Association for Computational Linguistics: {NAACL}
                  2025, Albuquerque, New Mexico, USA, April 29 - May 4, 2025},
  pages        = {4257--4275},
  year         = {2025}
}

@inproceedings{DBLP:conf/acl/GuiYYZSLC24,
  author       = {Honghao Gui and
                  Lin Yuan and
                  Hongbin Ye and
                  Ningyu Zhang and
                  Mengshu Sun and
                  Lei Liang and
                  Huajun Chen},
  editor       = {Lun{-}Wei Ku and
                  Andre Martins and
                  Vivek Srikumar},
  title        = {IEPile: Unearthing Large Scale Schema-Conditioned Information Extraction
                  Corpus},
  booktitle    = {Proceedings of the 62nd Annual Meeting of the Association for Computational
                  Linguistics, {ACL} 2024 - Short Papers, Bangkok, Thailand, August
                  11-16, 2024},
  pages        = {127--146},
  year         = {2024}
}

@article{DBLP:journals/corr/abs-2312-15548,
  author       = {Xinglin Xiao and
                  Yijie Wang and
                  Nan Xu and
                  Yuqi Wang and
                  Hanxuan Yang and
                  Minzheng Wang and
                  Yin Luo and
                  Lei Wang and
                  Wenji Mao and
                  Daniel Zeng},
  title        = {{YAYI-UIE:} {A} Chat-Enhanced Instruction Tuning Framework for Universal
                  Information Extraction},
  journal      = {CoRR},
  volume       = {abs/2312.15548},
  year         = {2023}
}

@inproceedings{DBLP:conf/emnlp/HanZYWYLS18,
  author       = {Xu Han and
                  Hao Zhu and
                  Pengfei Yu and
                  Ziyun Wang and
                  Yuan Yao and
                  Zhiyuan Liu and
                  Maosong Sun},
  editor       = {Ellen Riloff and
                  David Chiang and
                  Julia Hockenmaier and
                  Jun'ichi Tsujii},
  title        = {FewRel: {A} Large-Scale Supervised Few-shot Relation Classification
                  Dataset with State-of-the-Art Evaluation},
  booktitle    = {Proceedings of the 2018 Conference on Empirical Methods in Natural
                  Language Processing, Brussels, Belgium, October 31 - November 4, 2018},
  pages        = {4803--4809},
  year         = {2018}
}

@inproceedings{DBLP:conf/aaai/Liu0YDJCMF21,
  author       = {Zihan Liu and
                  Yan Xu and
                  Tiezheng Yu and
                  Wenliang Dai and
                  Ziwei Ji and
                  Samuel Cahyawijaya and
                  Andrea Madotto and
                  Pascale Fung},
  title        = {CrossNER: Evaluating Cross-Domain Named Entity Recognition},
  booktitle    = {Thirty-Fifth {AAAI} Conference on Artificial Intelligence, {AAAI}
                  2021, Thirty-Third Conference on Innovative Applications of Artificial
                  Intelligence, {IAAI} 2021, The Eleventh Symposium on Educational Advances
                  in Artificial Intelligence, {EAAI} 2021, Virtual Event, February 2-9,
                  2021},
  pages        = {13452--13460},
  year         = {2021}
}

@inproceedings{DBLP:conf/acl/ShengGYLHWLX21,
  author       = {Jiawei Sheng and
                  Shu Guo and
                  Bowen Yu and
                  Qian Li and
                  Yiming Hei and
                  Lihong Wang and
                  Tingwen Liu and
                  Hongbo Xu},
  editor       = {Chengqing Zong and
                  Fei Xia and
                  Wenjie Li and
                  Roberto Navigli},
  title        = {CasEE: {A} Joint Learning Framework with Cascade Decoding for Overlapping
                  Event Extraction},
  booktitle    = {Findings of the Association for Computational Linguistics: {ACL/IJCNLP}
                  2021, Online Event, August 1-6, 2021},
  series       = {Findings of {ACL}},
  volume       = {{ACL/IJCNLP} 2021},
  pages        = {164--174},
  year         = {2021}
}

@article{DBLP:journals/corr/abs-2412-20005,
  author       = {Yujie Luo and
                  Xiangyuan Ru and
                  Kangwei Liu and
                  Lin Yuan and
                  Mengshu Sun and
                  Ningyu Zhang and
                  Lei Liang and
                  Zhiqiang Zhang and
                  Jun Zhou and
                  Lanning Wei and
                  Da Zheng and
                  Haofen Wang and
                  Huajun Chen},
  title        = {OneKE: {A} Dockerized Schema-Guided {LLM} Agent-based Knowledge Extraction
                  System},
  journal      = {CoRR},
  volume       = {abs/2412.20005},
  year         = {2024}
}

@inproceedings{DBLP:conf/coling/LiaoDH025,
  author       = {Xincheng Liao and
                  Junwen Duan and
                  Yixi Huang and
                  Jianxin Wang},
  editor       = {Owen Rambow and
                  Leo Wanner and
                  Marianna Apidianaki and
                  Hend Al{-}Khalifa and
                  Barbara Di Eugenio and
                  Steven Schockaert},
  title        = {{RUIE:} Retrieval-based Unified Information Extraction using Large
                  Language Model},
  booktitle    = {Proceedings of the 31st International Conference on Computational
                  Linguistics, {COLING} 2025, Abu Dhabi, UAE, January 19-24, 2025},
  pages        = {9640--9655},
  year         = {2025}
}

@misc{yang2025qwen3technicalreport,
      title={Qwen3 Technical Report}, 
      author={An Yang and Anfeng Li and Baosong Yang and Beichen Zhang and Binyuan Hui and Bo Zheng and Bowen Yu and Chang Gao and Chengen Huang and Chenxu Lv and Chujie Zheng and Dayiheng Liu and Fan Zhou and Fei Huang and Feng Hu and Hao Ge and Haoran Wei and Huan Lin and Jialong Tang and Jian Yang and Jianhong Tu and Jianwei Zhang and Jianxin Yang and Jiaxi Yang and Jing Zhou and Jingren Zhou and Junyang Lin and Kai Dang and Keqin Bao and Kexin Yang and Le Yu and Lianghao Deng and Mei Li and Mingfeng Xue and Mingze Li and Pei Zhang and Peng Wang and Qin Zhu and Rui Men and Ruize Gao and Shixuan Liu and Shuang Luo and Tianhao Li and Tianyi Tang and Wenbiao Yin and Xingzhang Ren and Xinyu Wang and Xinyu Zhang and Xuancheng Ren and Yang Fan and Yang Su and Yichang Zhang and Yinger Zhang and Yu Wan and Yuqiong Liu and Zekun Wang and Zeyu Cui and Zhenru Zhang and Zhipeng Zhou and Zihan Qiu},
      year={2025},
      eprint={2505.09388},
      archivePrefix={arXiv},
      primaryClass={cs.CL},
      url={https://arxiv.org/abs/2505.09388}, 
}

@inproceedings{DBLP:conf/scie/Wilks97,
  author       = {Yorick Wilks},
  editor       = {Maria Teresa Pazienza},
  title        = {Information Extraction as a Core Language Technology},
  booktitle    = {Information Extraction: {A} Multidisciplinary Approach to an Emerging
                  Information Technology, International Summer School, SCIE-97, Frascati,
                  Italy, 14-18, 1997},
  series       = {Lecture Notes in Computer Science},
  volume       = {1299},
  pages        = {1--9},
  year         = {1997}
}

@inproceedings{DBLP:conf/naacl/ZaratianaTHC24,
  author       = {Urchade Zaratiana and
                  Nadi Tomeh and
                  Pierre Holat and
                  Thierry Charnois},
  editor       = {Kevin Duh and
                  Helena G{\'{o}}mez{-}Adorno and
                  Steven Bethard},
  title        = {GLiNER: Generalist Model for Named Entity Recognition using Bidirectional
                  Transformer},
  booktitle    = {Proceedings of the 2024 Conference of the North American Chapter of
                  the Association for Computational Linguistics: Human Language Technologies
                  (Volume 1: Long Papers), {NAACL} 2024, Mexico City, Mexico, June 16-21,
                  2024},
  pages        = {5364--5376},
  year         = {2024}
}

@inproceedings{DBLP:conf/acl/XuWLD0M23,
  author       = {Benfeng Xu and
                  Quan Wang and
                  Yajuan Lyu and
                  Dai Dai and
                  Yongdong Zhang and
                  Zhendong Mao},
  editor       = {Anna Rogers and
                  Jordan L. Boyd{-}Graber and
                  Naoaki Okazaki},
  title        = {S2ynRE: Two-stage Self-training with Synthetic data for Low-resource
                  Relation Extraction},
  booktitle    = {Proceedings of the 61st Annual Meeting of the Association for Computational
                  Linguistics (Volume 1: Long Papers), {ACL} 2023, Toronto, Canada,
                  July 9-14, 2023},
  pages        = {8186--8207},
  year         = {2023}
}

@article{DBLP:journals/corr/abs-2310-18463,
  author       = {Mingchen Li and
                  Ming Chen and
                  Huixue Zhou and
                  Rui Zhang},
  title        = {PeTailor: Improving Large Language Model by Tailored Chunk Scorer
                  in Biomedical Triple Extraction},
  journal      = {CoRR},
  volume       = {abs/2310.18463},
  year         = {2023}
}

@inproceedings{DBLP:conf/acl/ZhangBCSV24,
  author       = {Xinliang Frederick Zhang and
                  Carter Wood Blum and
                  Temma Choji and
                  Shalin Shah and
                  Alakananda Vempala},
  editor       = {Lun{-}Wei Ku and
                  Andre Martins and
                  Vivek Srikumar},
  title        = {{ULTRA:} Unleash LLMs' Potential for Event Argument Extraction through
                  Hierarchical Modeling and Pair-wise Self-Refinement},
  booktitle    = {Findings of the Association for Computational Linguistics, {ACL} 2024,
                  Bangkok, Thailand and virtual meeting, August 11-16, 2024},
  pages        = {8172--8185},
  year         = {2024}
}

@inproceedings{DBLP:conf/coling/FanL0YHL24,
  author       = {Yuchen Fan and
                  Yantao Liu and
                  Zijun Yao and
                  Jifan Yu and
                  Lei Hou and
                  Juanzi Li},
  editor       = {Nicoletta Calzolari and
                  Min{-}Yen Kan and
                  V{\'{e}}ronique Hoste and
                  Alessandro Lenci and
                  Sakriani Sakti and
                  Nianwen Xue},
  title        = {Evaluating Generative Language Models in Information Extraction as
                  Subjective Question Correction},
  booktitle    = {Proceedings of the 2024 Joint International Conference on Computational
                  Linguistics, Language Resources and Evaluation, {LREC/COLING} 2024,
                  20-25 May, 2024, Torino, Italy},
  pages        = {6409--6417},
  year         = {2024}
}

@inproceedings{DBLP:conf/ccks/ZhaoWK22,
  author       = {Fubang Zhao and
                  Yexiang Wang and
                  Yangyang Kang},
  editor       = {Ningyu Zhang and
                  Meng Wang and
                  Tianxing Wu and
                  Wei Hu and
                  Shumin Deng},
  title        = {A Prompt-Based {UIE} Framework},
  booktitle    = {{CCKS} 2022 - Evaluation Track - 7th China Conference on Knowledge
                  Graph and Semantic Computing Evaluations, {CCKS} 2022, Qinhuangdao,
                  China, August 24-27, 2022, Revised Selected Papers},
  series       = {Communications in Computer and Information Science},
  volume       = {1711},
  pages        = {163--171},
  year         = {2022}
}

@article{DBLP:journals/corr/abs-2304-08085,
  author       = {Xiao Wang and
                  Weikang Zhou and
                  Can Zu and
                  Han Xia and
                  Tianze Chen and
                  Yuansen Zhang and
                  Rui Zheng and
                  Junjie Ye and
                  Qi Zhang and
                  Tao Gui and
                  Jihua Kang and
                  Jingsheng Yang and
                  Siyuan Li and
                  Chunsai Du},
  title        = {InstructUIE: Multi-task Instruction Tuning for Unified Information
                  Extraction},
  journal      = {CoRR},
  volume       = {abs/2304.08085},
  year         = {2023}
}

@inproceedings{DBLP:conf/naacl/DevlinCLT19,
  author       = {Jacob Devlin and
                  Ming{-}Wei Chang and
                  Kenton Lee and
                  Kristina Toutanova},
  editor       = {Jill Burstein and
                  Christy Doran and
                  Thamar Solorio},
  title        = {{BERT:} Pre-training of Deep Bidirectional Transformers for Language
                  Understanding},
  booktitle    = {Proceedings of the 2019 Conference of the North American Chapter of
                  the Association for Computational Linguistics: Human Language Technologies,
                  {NAACL-HLT} 2019, Minneapolis, MN, USA, June 2-7, 2019, Volume 1 (Long
                  and Short Papers)},
  pages        = {4171--4186},
  year         = {2019}
}

@article{DBLP:journals/ml/CortesV95,
  author       = {Corinna Cortes and
                  Vladimir Vapnik},
  title        = {Support-Vector Networks},
  journal      = {Mach. Learn.},
  volume       = {20},
  number       = {3},
  pages        = {273--297},
  year         = {1995}
}

@misc{deepseekai2025deepseekr1incentivizingreasoningcapability,
      title={DeepSeek-R1: Incentivizing Reasoning Capability in LLMs via Reinforcement Learning}, 
      author={DeepSeek-AI},
      year={2025},
      eprint={2501.12948},
      archivePrefix={arXiv},
      primaryClass={cs.CL},
      url={https://arxiv.org/abs/2501.12948}, 
}

@techreport{Meta2024Llama,
  author = {{Meta AI}},
  title = {Llama 3.1 Technical Report: Language, Vision, and Multimodal Capabilities},
  institution = {Meta},
  year = {2024},
  url = {https://llama.meta.com/},
  note = {Model parameters: 405B, context length 128K}
}

@article{chunkuie2025,
  author       = {Li, Wei and Liu, Yingzhen and Yang, Yinling and Zhang, Ting and Men, Wei},
  title        = {ChunkUIE: Chunked instruction-based unified information extraction},
  journal      = {PLOS ONE},
  year         = {2025},
  volume       = {20},
  number       = {6},
  pages        = {e0326764},
  doi          = {10.1371/journal.pone.0326764},
  publisher    = {Public Library of Science (PLoS)},
  issn         = {1932-6203},
}

@inproceedings{Chiticariu2013RuleBasedIE,
  author = {Laura Chiticariu and Yunyao Li and Frederick R. Reiss},
  title = {Rule‑Based Information Extraction is Dead! Long Live Rule‑Based Information Extraction Systems!},
  booktitle = {Proceedings of EMNLP},
  year = {2013},
  pages = {827--832},
  address = {Seattle, Washington, USA},
  publisher = {Association for Computational Linguistics}
}

@article{Maturana2017DocumentSpanners,
  author = {Francisco Maturana and Cristian Riveros and Domagoj Vrgoč},
  title = {Document Spanners for Extracting Incomplete Information: Expressiveness and Complexity},
  journal = {arXiv preprint arXiv:1707.00827},
  year = {2017}
}

@incollection{Thenmozhi2018RCEOIE,
  author = {D. Thenmozhi and Chandrabose Aravindan},
  title = {RCE‑OIE: Open Information Extraction Using a Rule‑Based Clause Extraction Engine},
  booktitle = {Recent Findings in Intelligent Computing Techniques},
  series = {Advances in Intelligent Systems and Computing},
  volume = {709},
  year = {2018},
  pages = {191--198},
  publisher = {Springer}
}

@article{DBLP:journals/corr/abs-2505-11739,
  author       = {Feijiang Han and
                  Xiaodong Yu and
                  Jianheng Tang and
                  Lyle H. Ungar},
  title        = {ZeroTuning: Unlocking the Initial Token's Power to Enhance Large
                  Language Models Without Training},
  journal      = {CoRR},
  volume       = {abs/2505.11739},
  year         = {2025},
  url          = {https://doi.org/10.48550/arXiv.2505.11739},
  doi          = {10.48550/ARXIV.2505.11739},
  eprinttype    = {arXiv},
  eprint       = {2505.11739},
  bibsource    = {dblp computer science bibliography, https://dblp.org}
}

\end{document}